%% file: main.tex
\documentclass[10pt,twocolumn,letterpaper]{article}

\usepackage[table, dvipsnames]{xcolor}
\definecolor{rowblue}{RGB}{220,230,240}
\usepackage{iccv}
\usepackage{times}
\usepackage{epsfig}
\usepackage{graphicx}
\usepackage{amsmath}
\usepackage{amssymb}
\usepackage{algorithm}
\usepackage{algorithmic}
\usepackage{xcolor}
\usepackage{ctable} 
\usepackage{pifont}

\usepackage{multirow}
\usepackage[group-separator={,}, group-minimum-digits=4]{siunitx}

\usepackage[pagebackref=true,breaklinks=true,letterpaper=true,colorlinks,bookmarks=false]{hyperref}

\iccvfinalcopy 

\def\iccvPaperID{1343} 

\ificcvfinal\fi
\begin{document}

\title{Anchor Diffusion for Unsupervised Video Object Segmentation}
\author{Zhao Yang\thanks{Equal contribution.}\\
University of Oxford\\
{\tt\small zhao.yang@eng.ox.ac.uk}
\and
Qiang Wang$^*$\\
CASIA\\
{\tt\small qiang.wang@nlpr.ia.ac.cn}
\and
Luca Bertinetto\\
Five AI\\
{\tt\small luca@robots.ox.ac.uk}
\and
Weiming Hu\\
CASIA\\
{\tt\small wmhu@nlpr.ia.ac.cn}
\and
Song Bai\\
University of Oxford\\
{\tt\small songbai.site@gmail.com}
\and
Philip H.S. Torr\\
University of Oxford\\
{\tt\small philip.torr@eng.ox.ac.uk}
}
\maketitle
\input{sections/abstract}

\input{sections/introduction}
\input{sections/related_work}
\input{sections/method}

\input{sections/experiments}

\input{sections/conclusion}
{\small
\bibliographystyle{ieee_fullname}
\bibliography{main}
}

\input{sections/supplementary}

\end{document}

%% file: sections/abstract.tex
\begin{abstract}
Unsupervised video object segmentation has often been tackled by methods based on recurrent neural networks and optical flow.
Despite their complexity, these kinds of approaches tend to favour short-term temporal dependencies and are thus prone to accumulating inaccuracies, which cause drift over time.
Moreover, simple (static) image segmentation models, alone, can perform competitively against these methods, which further suggests that the way temporal dependencies are modelled should be reconsidered.
Motivated by these observations, in this paper we explore simple yet effective strategies to model long-term temporal dependencies.
Inspired by the non-local operators of~\cite{non-local}, we introduce a technique to establish dense correspondences between pixel embeddings of a reference ``anchor'' frame and the current one.
This allows the learning of pairwise dependencies at arbitrarily long distances without conditioning on intermediate frames.
Without online supervision, our approach can suppress the background and precisely segment the foreground object even in challenging scenarios, while maintaining consistent performance over time.
With a mean IoU of $81.7\%$, our method ranks first on the DAVIS-2016 leaderboard of unsupervised methods, while still being competitive against state-of-the-art online semi-supervised approaches.
We further evaluate our method on the FBMS dataset and the ViSal video saliency dataset, showing results competitive with the state of the art.
\end{abstract}

%% file: sections/introduction.tex
\section{Introduction} \label{sec:introduction}
\begin{figure*}[tb]
\centering
\includegraphics[width=0.99\linewidth]{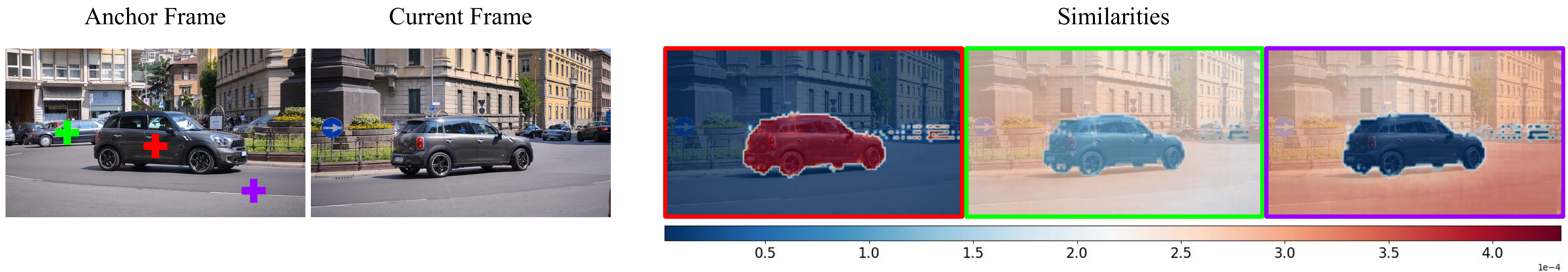}
\caption{
Example of one-to-many similarities (right-hand side) learned by our method between single pixel embeddings belonging to the \emph{anchor frame} to all the pixel embeddings from the \emph{current frame} (left-hand side).
The outlines of the dense similarities match the colour of the corresponding pixel embedding in the anchor frame.
Notice how the dense similarities with the pixel embedding from the foreground car (in red) produce a neat heat map that well identifies the object, while both sets of similarities with the pixel embedding from a ``distractor'' car (in green) and the road (in purple) are higher in correspondence of the background.
These learned similarities are a simple and effective way of segmenting out foreground objects.
Best viewed in colours.
}
\label{fig:heatmap}
\end{figure*}
Video object segmentation (VOS) is a fundamental task in many important areas such as autonomous driving~\cite{carla, kitti, oxfordrobotcar}, robotic manipulation~\cite{robot}, video surveillance~\cite{surv} and video editing~\cite{perazzi2017video}.
Contemporary literature typically considers this problem in either the \emph{semi-supervised} or the \emph{unsupervised} setting.
In both cases the objective is to predict, in every frame, pixel-level masks delineating certain objects of interest.

Under the semi-supervised setting, at test time methods can rely on a mask that specifies the object to segment.
In contrast, the unsupervised setting does not provide any initialisation.
Without online supervision, the task might be considered ambiguous, as different objects could be considered of interest for different reasons, according to the application.
Among researchers, the current consensus is to segment foreground objects where a human gaze is more likely to focus~\cite{davis2019}.
In more practical terms, an object is generally considered as \emph{foreground} if it is sufficiently large, in motion and centred in the scene.
In certain datasets (\eg,~FBMS~\cite{fbms} and ViSal~\cite{visal}), in the same video, multiple foreground objects are considered, while in DAVIS-2016~\cite{davis} only a single object is considered.

With the aim of tracking temporal changes of target objects, current state-of-the-art unsupervised approaches generally model motion cues in a video sequence via optical flow~\cite{segflow, fusionseg, arp, iet, mbn, teacher-student, mp, gru-mp} or recurrent neural networks (RNNs)~\cite{lstm,iet,gru-mp}.
Typically, these methods sequentially propagate features from the previous steps to the current one, thus making the current prediction depending on the entire history of the video.

Though having the potential of exploiting informative temporal cues, these approaches suffer from several limitations.
RNNs often rely on training techniques such as truncated backpropagation through time to reduce the cost of parameter updates, which limits their long-term modelling capability~\cite{sutskever}.
Moreover, while LSTM's gating mechanism alleviates the issue of vanishing gradients~\cite{bengio,difficultRNN}, the phenomenon of exploding gradients often requires clipping or rescaling the norm of the gradients during training~\cite{seq2seq}.
Optical flow vectors only predict one-step motion cues at each frame in a video, which can accumulate errors over time.
What is more, models relying on optical flow are typically trained on synthetic videos due to the high cost of per-frame and per-pixel labelling.
Therefore, when applying these systems to real videos, the domain gap can cause the flow fields to contain several inaccuracies, especially when the foreground is nearly static~\cite{gru-mp}.

In the video object segmentation community, the deterioration of performance over time in unsupervised VOS methods based on optical flow or RNNs is well known and has been widely discussed~\cite{pml, iet, rgmp, onavos}.
For instance, Li~\etal~\cite{iet} demonstrate that, as a regular optical flow-based model progresses through frames, foreground embeddings become increasingly closer in feature space to the first frame's background as opposed to the foreground.
Furthermore, Voigtlaender~\etal~\cite{onavos} observe that a simple static segmentation model can achieve competitive results in the unsupervised VOS setting, which further corroborates the case for steering away from the sequential modelling strategies used by established methods.

Motivated by the above observations, in this work we opt for a much simpler solution, which is based on learning the similarity between pixels belonging to frames that can be arbitrarily far apart in time.
To ensure representation consistency and reduce long-term drift, we propagate the features of the first frame (the ``anchor'') to the current frame via an aggregation technique inspired by the non-local operation introduced by Wang~\etal~\cite{non-local}.
This approach allows us to forgo of sequential modelling, while at the same time enabling us to deal with long-term dependencies and achieve high robustness over time, as shown in our experiments.

Despite its simplicity and online operability, our method outperforms the current state of the art~\cite{teacher-student} on the DAVIS-2016 leaderboard by a margin of (absolute) $2.2\%$ in terms of intersection-over-union, without resorting to auxiliary training data or post-processing.
Moreover, it also achieves state-of-the-art results on FBMS~\cite{fbms} and the ViSal~\cite{visal} video saliency benchmark.
Code and pre-trained models are available at~\url{https://github.com/yz93/anchor-diff-VOS}.

%% file: sections/related_work.tex
\section{Related work} \label{sec: related-work}
The problem of video object segmentation (VOS) is tackled by the computer vision community in the \emph{unsupervised} or \emph{semi-supervised} settings, which are defined by the level of supervision provided at test time.

\vspace{1ex}\noindent\textbf{Semi-supervised VOS} methods are provided with a pixel-wise mask identifying the target object in the first frame of a video.
When aiming at very high segmentation accuracy, methods generally perform online fine-tuning on the basis of this supervision~\cite{osvos,lucid,vsreid,premvos,osvoss,msk,onavos}, sometimes exploiting data-augmentation techniques~\cite{osvos,lucid} or self-supervision~\cite{onavos}.
As online fine-tuning can take up to several minutes per video, many recently proposed methods renounce to it and instead aim at a faster online speed (\eg,~\cite{pml,favos,siammask}).
These faster semi-supervised approaches come in many flavours.
For instance, Chen~\etal~\cite{pml} learn a metric space for pixel embeddings, which is then used to establish associations between pixels across frames, while Cheng~\etal~\cite{favos} suggest to individually track object parts from the first frame with a visual object tracker~\cite{siamFC} and then aggregate them according to their similarity with the initialisation mask.

\vspace{1ex}\noindent\textbf{Unsupervised VOS} methods, instead, cannot rely on any supervision at test time and are often based on optical flow and RNNs.
The purely optical flow-based MP-Net~\cite{mp} discards appearance modelling and casts segmentation as foreground motion prediction, an approach which poorly deals with static foreground objects.
To address this problem, several methods (\eg,~LVO~\cite{gru-mp}, SegFlow~\cite{segflow}, MotAdapt~\cite{teacher-student} and MBN~\cite{mbn}) suggest to integrate appearance-based and optical flow-based features together, leading to variations of the ``two-stream model'' presenting two dedicated parallel branches.
The drawbacks of these methods are threefold.
First, flow estimation networks are typically trained on synthetic datasets and can thus result in poor performance when deployed in the real world.
Second, while modelling long-term temporal dependencies is critical for adapting to significant online changes, the vector fields can only model short-term one-step dependencies.
Targeting this issue, Tokmakow~\etal~\cite{gru-mp} proposed to extend the horizon spanned by optical flow-based features by employing a convolutional gated recurrent unit~\cite{gru}.
Third, vector fields cannot distinguish foreground and background objects when they move in a synchronised fashion (\eg,~the cars in a traffic jam).
Li~\etal~\cite{mbn} attempt to address this issue by employing a bilateral network for detecting the motion of background objects.
Our investigations with a much simpler appearance-based approach show that optical flow may not be an essential component of unsupervised VOS systems.

RNN-based models are often challenged by the problems of exploding and vanishing gradients~\cite{bengio,difficultRNN}, which limit their long-term modelling capability.
Among the methods that make use of recurrent connections, Song~\etal~\cite{pdb} propose a novel convolutional long short-term memory~\cite{lstm} architecture, in which two atrous convolution~\cite{deeplabv3} layers are stacked along the forward axis and propagate features in opposite directions.

Recently, it has been shown~\cite{iet,rgmp,onavos} that both recurrent and optical flow-based methods significantly suffer from a deterioration in the quality of their predictions over time.
This has motivated the several approaches (including ours) that tackle video object segmentation by simply learning similarities between pixel embeddings (\eg,~\cite{pml,instanceDML,iet,mbn}).
These methods first select a set of seed pixels that are most likely to belong to the foreground object and then classify all other pixels based on their similarities to these seeds, for instance by thresholding or by propagating labels between neighbours.
Fathi~\etal~\cite{instanceDML} adopt this approach for semantic instance segmentation, in which the pairwise pixel similarity function measures the likelihood of two pixels belonging to the same instance.
IET~\cite{iet} extends this concept to video sequences.
Similarly, it selects a set of foreground and background seeds for each frame and organises them into tracks.
It then segments each frame individually based on pixel similarities with the foreground and background seeds.
Note that IET utilises pre-trained instance embeddings.
MBN~\cite{mbn} extends IET with a bilateral filtering network that filters false-positive foreground predictions using optical flow features and an energy minimisation procedure on a graph of seeds sampled from a few consecutive frames.
When segmenting frame $t$, MBN classifies each pixel by assigning it the label of the seed (sampled from frames $t{-}1$, $t$, and $t{+}1$) with which it has the smallest embedding distance.

The main drawback of these methods is in the complexity involved in the procedures of seed selection, ranking and classification, critical for achieving good performance.
Moreover, these algorithms also depend on multiple scores such as motion saliency and objectness that need to be carefully calibrated and combined into one final metric.

Albeit our proposal is related to this last class of approaches, it is considerably simpler.
Instead of separately learning individual components from image datasets and classifying pixels based on similarities with seeds, our method performs similarity learning, feature propagation and binary segmentation in a single network.

%% file: sections/method.tex
\section{Method} \label{sec: method}
\begin{figure*}[t]
  \centering
  \includegraphics[width=0.99\linewidth]{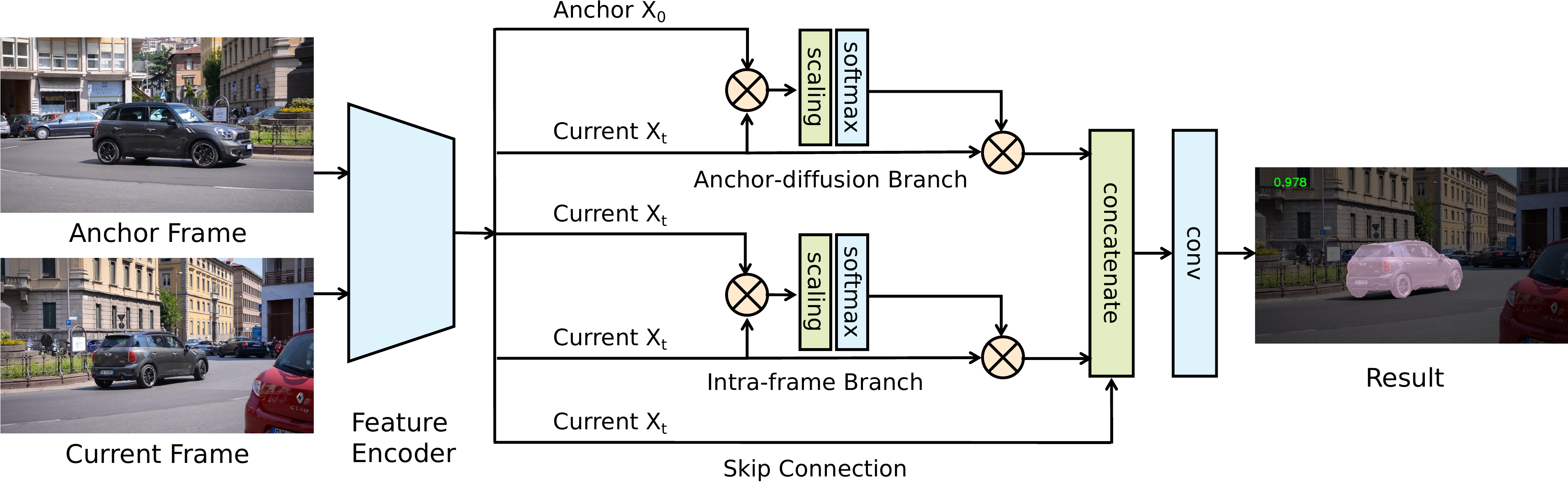}
  \caption{Overall pipeline of the proposed method.
		In the anchor-diffusion branch, pixel embeddings in the current frame are linearly transformed by similarity scores with pixel embeddings in the anchor frame, and concatenated with both the output of the intra-frame branch and the output of the skip connection.}
  \label{fig:pipeline}
\end{figure*}

We are interested in the task of binary segmentation of a sequence of video frames, where the final performance is measured by the average segmentation quality of individual frames.
Therefore, our method should perform well under two aspects.
First, similarly to what is expected from a static segmentation model, it should be able to provide accurate segmentation masks of foreground objects in individual frames.
Second, it should be able to well adapt to the appearance changes of the foreground objects throughout the whole video.

In the proposed \emph{anchor diffusion network} (AD-Net) (Figure~\ref{fig:pipeline}), we address both requirements in a single model trained end-to-end by leveraging the recently proposed non-local operations of Wang \etal~\cite{non-local}.
Closely related to the concept of self-attention~\cite{attention-all-you-need}, a non-local operation is a neural network building block that captures the dependencies within a set of input feature vectors.

To achieve our first goal, a non-local operation is applied to the encoding of the target frame, in a similar way it is applied for semantic image segmentation~\cite{dual-attention}, forming the \emph{intra-frame branch} of our overall model.
To achieve our second goal, we propagate information between two frames: a fixed anchor frame and the current frame, forming the \emph{anchor-diffusion branch} of our overall model.
We name the branch this way to give relevance to its functionality of ``diffusing'' information from the anchor to the large number of target frames at test time, which encourages foreground embeddings of each target frame to be consistent over time.

In the following, we describe our pipeline in more detail.

\vspace{1ex}\noindent\textbf{Pipeline}.
The input of our model consists of a pair of images: a fixed anchor frame $I_0$  and the frame to segment $I_t$.
The overall pipeline is schematically illustrated in Figure~\ref{fig:pipeline}.
First, a feature encoder (the fully-convolutional DeepLabv3~\cite{deeplabv3}) encodes $I_0$ and $I_t$ into the corresponding embeddings $X_0\in\mathbb{R}^{hw\times c}$ and $X_t\in\mathbb{R}^{hw\times c}$, where $c$ denotes the number of channels and $h$, $w$ denote the height and width of the frame.
We refer to the $c$-dimensional feature vector at each location as a \emph{pixel embedding}.
The output of this first stage is then fed to three parallel branches: a skip connection with an identity mapping~\cite{resnet}, the intra-frame branch, and the anchor-diffusion branch.
$X_t$ is fed to all branches, while $X_0$ only to the anchor-diffusion branch.
Finally, the resulting features from the three branches are concatenated together along the channel dimension before the classification layer.

The entire network is trained end-to-end with a binary cross-entropy loss.
Though any frame could be selected as the anchor frame, in practice we always choose the first frame for computational convenience and because, in benchmarks, the first frame is guaranteed to contain the foreground objects.
During training, the first frame and a random frame are sampled from the video.

\subsection{Anchor diffusion} \label{sec:ad-net}
As described earlier, $X_0$ and $X_t$ represent the embeddings of the anchor and the current frame respectively.
To reinforce the foreground signal, it is important to know which pixel embeddings in $X_t$ correspond to the background introduced throughout a video.
To achieve this, in the anchor-diffusion branch we compute a transition matrix $P\in\mathbb{R}^{hw\times hw}$ which establishes dense correspondences between each pair of pixels from $X_0$ and $X_t$ and use it to map $X_t$ to a new encoding $\tilde{X_t}$, in which the pixel embeddings are weighted according to their similarity with the foreground:
\begin{equation} \label{eq:1}
\tilde{X_t}=PX_t.
\end{equation}
As qualitatively illustrated in Figure~\ref{fig:heatmap} and Appendix~\ref{sec:correlation}, this procedure significantly strengthens the foreground while weakening the background.
It is worth noting that one can also simply use the concatenation of $X_0$ and $X_t$ to achieve this goal.
However, we find in our experiments that the correspondence learning in Equation~\eqref{eq:1} can better localise the foreground objects.

Similarly to~\cite{non-local}, the transition matrix is defined as
\begin{equation} \label{eq:2}
P=\mbox{softmax}(\frac{1}{z}X_0 X_t^T),
\end{equation}
where $X_0 X_t^T$ is a pairwise dot product similarity between each pair of pixel embeddings in $X_0$ and $X_t$.
Following~\cite{bilinear,attention-all-you-need}, we scale the dot product with a factor $z=\sqrt{c}$, where $c$ is the number of channels of $X_0$ and $X_t$.
The rationale being that, for embeddings with high dimensionality, dot products can be very large and thus push the output of the softmax to regions where gradients are small~\cite{attention-all-you-need}.
The softmax function normalises each row of $\frac{1}{z}X_0 X_t^T$ to sum to one, thereby preserving scale invariance of the pixel embeddings.
Without normalisation, multiplying $\frac{1}{z}X_0 X_t^T$  with $X_t$ can entirely change the scale of the pixel embeddings.

In the case of the intra-frame branch, each output pixel embedding can be considered as a global aggregation of all input pixel embeddings weighted by pairwise appearance similarity.
It has been shown that such use of non-local operations~\cite{non-local} can harness long-range spatial information, which is beneficial for semantic segmentation~\cite{dual-attention}.
Empirically, as detailed in the ablation studies of Table~\ref{tab:AD-Net}, we found that incorporating this branch in addition to the anchor-diffusion branch further improves the performance of the model.

The intra-frame branch improves segmentation accuracy but does not address the temporal changes in a video sequence.
Conversely, the anchor-diffusion branch models pairwise dependencies between frames, with the result of enhancing the consistency of pixel embeddings and reducing drift.

\vspace{1ex}\noindent\textbf{Qualitative analysis}.
As shown in Figure~\ref{fig:heatmap}, each of the coloured pixels in the anchor frame finds desirable correspondences in the current frame.
The foreground car pixel embedding (red) has high similarity with pixel embeddings of the foreground car in the current frame despite the appearance change and sets off a neat contrast with the background that precisely outlines the target object.
Conversely, both heat maps of the distractor car pixel embedding (green) and the road pixel embedding (purple) have higher similarity values in the background region of the current frame, which is what expected for pixel embeddings of the background class.
Moreover, as the distractor car is not present in the current frame, its pixel embedding only find weak and widespread correspondences in the general background region, with a weak separation between the foreground and the background.
In contrast, the pixel embedding corresponding to the asphalt, which represents the common material appearing in both frames, shows a higher similarity with the road region and sets a larger separation between the foreground and the background.
More qualitative results illustrating the similarity between pixel embeddings are showed in Appendix~\ref{sec:correlation}.

Overall, these results show that the transition matrix $P$ learns a similarity metric that can well identify common objects/materials across two frames.
Therefore, when used in Equation~\eqref{eq:1}, $P$ can strengthen the signal from pixels which have strong correspondences in the anchor frame and weaken the signal from pixels which do not.
As the foreground target object is almost always present in both frames while the background changes relatively quickly, our diffusion process generally strengthens the foreground and suppresses the background.

\begin{figure}[tb]
\centering
\includegraphics[width=0.49\textwidth]{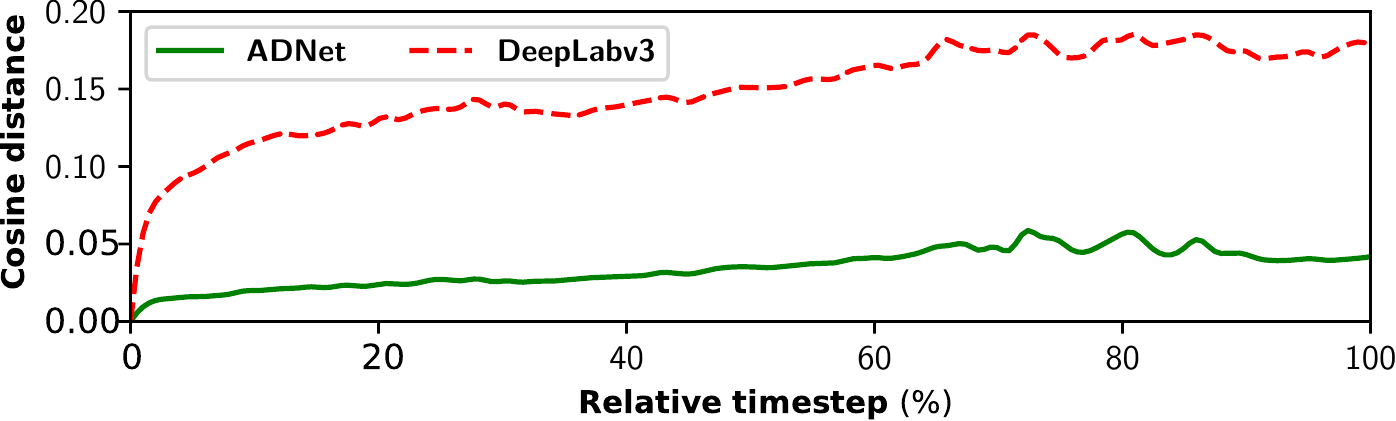}
\caption{Temporal consistency of pixel embeddings over time, measured as cosine distance between foreground pixel embeddings in the anchor frame and foreground pixel embeddings in progressively more distant frames.}
\label{fig:embeddings}
\end{figure}

In Figure~\ref{fig:embeddings}, instead, we report how foreground embeddings change over time by computing the average cosine distance between the foreground embeddings of a later frame and those of the first frame.
Notice how the embeddings of the baseline quickly grow apart, while the ones learned with our proposed method are significantly stabler.
This suggests that AD-Net is capable of preserving the foreground information from the first frame in a video over long time-frames.

%% file: sections/experiments.tex
\section{Experiments} \label{sec:experiments}
In the following, after discussing important implementation details regarding our architecture and training procedure, in Section~\ref{sec:benchmarks} we illustrate the three benchmarks we adopted, in Section~\ref{sec:ablation} we describe several ablation studies and in Section~\ref{sec:comparison}, we provide an extensive comparison with the state of the art.

\vspace{1ex}\noindent\textbf{Implementation details}.
We employ the fully-convolutional DeepLabv3~\cite{deeplabv3} as the feature encoder, and initialise its ResNet101~\cite{resnet} backbone with weights pre-trained on ImageNet.
The other layers in DeepLabv3 are randomly initialised.
The configuration of the dilation rates follows the original model~\cite{deeplabv3} and presents a total stride of $8$.
We modify the number of output channels in the last layer to $128$, which corresponds to $c$ in Section~\ref{sec: method}.

In the anchor-diffusion step, the spatial dimensions of each image encoding are flattened and transposed where appropriate in order to perform batched matrix multiplication.
The outputs of the three branches are concatenated along the channel axis and reduced to dimension $128$ via a $1{\times}1$ convolution with LeakyReLU non-linearity and dropout rate $0.1$.
The final classification layer is implemented as a $1{\times}1$ convolution with a single output channel followed by a sigmoid layer.

\vspace{1ex}\noindent\textbf{Training}.
Each training example consists of a pair of images.
Given a randomly sampled video, we use the first frame as the anchor image and a randomly sampled frame as the second image.
We also experimented with randomly sampling both frames and observed slightly worse performance.
Each input frame is cropped to a randomly-sized region enclosing the ground-truth foreground.
Random rotations are performed at $45$-degree increments, with a probability of $51\%$ of not rotating and equal probabilities of rotating to any of the remaining angles.

The model is trained with binary cross-entropy loss.
Network parameters are optimised via stochastic gradient descent with a weight decay of $0.0005$.
The initial learning rate is set to $0.005$ and follows a ``poly'' adjustment policy~\cite{deeplabv3}, where the initial learning rate is multiplied by $(1-\frac{iter}{40,000})^{0.9}$ at each iteration.
The model is trained for \num{30000} iterations with batch size $8$.
Raw predictions are upsampled via bilinear interpolation to the size of the ground-truth masks.

\vspace{1ex}\noindent\textbf{Inference}.
At test time, the features of the anchor frame are computed once and reused throughout the video.
Multi-scale and mirrored inputs are employed to enhance the final performance.
Each input image is scaled by factors of $0.75$, $1.00$ and $1.50$ and horizontally flipped.
The final heatmap is the mean of all output heatmaps.
Thresholding at $0.5$ produces the final binary labels.

\vspace{1ex}\noindent\textbf{Instance pruning}.
Since semantic segmentation approaches like the one we use lack the notion of instance and some videos from the DAVIS-2016 dataset~\cite{davis} present multiple objects that can be deemed as foreground, we experiment with a simple set of post-processing steps to prune ``non-foreground'' objects.
As instance trajectories measure the spatial changes of an instance, they can be used to detect background instances which have distinct trajectory patterns than the foreground instance.  
First, we establish online temporal correspondences by using a pre-trained object detection model~\cite{extremenet} to predict the locations of all objects and track the trajectory of each detection across the entire video using an intersection-over-union criterion between consecutive bounding boxes.
Once object tracks have been established, we use the cumulative area of instance masks across frames as a proxy to identify foreground objects, thus pruning small objects or objects that are only present in a fraction of the video.
This process produces a filtering mask, which is multiplied element-wise with AD-Net predictions to obtain the final predictions.
More details and hyper-parameters related to this process (which we refer to as \emph{instance pruning}) are provided in Appendix~\ref{sec:instance-pruning}.

\subsection{Benchmarks} \label{sec:benchmarks}
\noindent\textbf{Datasets}.
DAVIS~\cite{davis} is a benchmark and yearly challenge for video object segmentation (VOS).
Unsupervised methods are trained and evaluated with the DAVIS-2016 dataset, which annotates a single foreground entity.
There are 30 videos for training and 20 videos for validation.
We train our method on the training set and evaluate on the validation set.

The FBMS~\cite{fbms} dataset is another challenging benchmark for unsupervised video object segmentation containing $29$ training videos and $30$ test videos.
Following~\cite{iet,mbn,teacher-student,pdb,gru-mp}, we evaluate on the test set.

Finally, ViSal~\cite{visal} is a video salient object detection dataset containing $17$ video sequences.
Despite our method has not been designed for the task of saliency, we can easily report results on this benchmark too.

\vspace{1ex}\noindent\textbf{Evaluation metrics}.
For DAVIS, we adopt the official evaluation metrics of mean region similarity $\mathcal{J}$, which is the intersection-over-union of the prediction and ground truth, and mean contour accuracy $\mathcal{F}$, which is the F-measure defined on contour points from the prediction and the ground truth.
To provide more insights, we plot precision-recall (PR) curves on all three benchmark datasets.
On the FBMS dataset, the main evaluation metric is the F-measure.
On the ViSal dataset, we report the mean absolute error (MAE) and the F-measure.
For definitions of MAE and the F-measure, we refer readers to~\cite{dss}.

\subsection{Ablation studies} \label{sec:ablation}
\begin{table}[tb]
\begin{center}
\small
\begin{tabular}{l|cr|cr}
Model & ${\mathcal{J}}${\footnotesize (\%)} &  $\vartriangle_\mathcal{J}$ & ${\mathcal{F}}${\footnotesize (\%)} &  $\vartriangle_\mathcal{F}$  \\
\specialrule{1pt}{1pt}{1pt}
Baseline~\cite{deeplabv3}  & 75.41 & 0.00 & 75.58 & 0.00  \\
Baseline + intra-frame    & 76.17 & +0.76 & 75.38 & -0.20  \\
Baseline + anchor    & 76.84 & +1.43 & 75.76 & +0.18  \\
Baseline + anchor-diffusion &77.43 &+2.02 &76.78 &+1.20 \\
\specialrule{0.5pt}{0.5pt}{0.5pt}
AD-Net (single scale) & 78.26 &+2.85 &77.11 &+1.53 \\ \hline
\end{tabular}
\end{center}
\caption{Ablation study on the DAVIS-2016 validation set.
$\vartriangle_\mathcal{J}$ and $\vartriangle_\mathcal{F}$ denote, respectively, absolute improvements in region similarity and contour accuracy.}
\label{tab:AD-Net}
\end{table}

We conduct several ablations to evaluate the effectiveness of the anchor-diffusion procedure.
First, we evaluate DeepLabv3~\cite{deeplabv3+} \emph{as-is}, simply fine-tuning it on the DAVIS training set.
This semantic segmentation baseline (designed for static images) performs on par with some state-of-the-art unsupervised VOS methods (see Table~\ref{table:davis-2016}).
This is in line with what described by Voigtlaender~\etal~\cite{onavos}, but it is rather curious that it still applies after two years of progress.
Clearly, the competitive performance can be partially attributed to the high performance of DeepLabv3 for the similar task of semantic segmentation of static images.
However, this result also shows that existing unsupervised VOS techniques are not able to successfully model and leverage temporal dependencies and that different approaches should be sought.

Starting from this baseline, we evaluate four variants that differ in the embeddings they consider at the terminal concatenation layer (see Figure~\ref{fig:pipeline}).
Each corresponds to a row below \emph{Baseline} in Table~\ref{tab:AD-Net}.
The first variant (``intra-frame'') computes non-local features within the same frame $X_t$ and without the anchor-diffusion branch.
The second (``anchor'') simply concatenates $X_0$ to $X_t$.
The third performs anchor diffusion on $X_0$ and $X_t$, and concatenates the results with $X_t$, without features from the intra-frame branch.
The fourth (our final model, AD-Net) concatenates both the output of the intra-frame branch and that of the anchor-diffusion branch with $X_t$.

The ``intra-frame'' variant improves over the baseline, which shows the potential of utilising context information within the current frame.
The ``anchor'' variant demonstrates the general usefulness of an anchor frame, despite the apparent limitation that the fixed representation of the anchor frame does not adapt to changes in the current frame.
The solid performance gains validate our motivation to further develop the anchor-diffusion mechanism.
The ``anchor-diffusion'' variant illustrates the efficacy of the proposed feature diffusion mechanism across the anchor and current frames.
It brings a performance boost of $2.02$ (absolute) points over the baseline, larger than the contribution brought by the ``intra-frame" and ``anchor" variants.

\subsection{Comparison with the state of the art}\label{sec:comparison}
\begin{table}[t]
    \centering
    \small
    \resizebox{\columnwidth}{!}
    {
	\begin{tabular}{c|r|ccc|cc|c}
	\specialrule{1pt}{1pt}{1pt}
	    &&&&& \multicolumn{2}{c|}{DAVIS} & FBMS \\
		 & Method & FF & OF & CRF & ${\mathcal{J}}$ & ${\mathcal{F}}$ & F-measure \\ \hline
		\specialrule{1pt}{1pt}{1pt}
		\multirow{6}{*}{\begin{tabular}[r]{@{}c@{}}\rotatebox{90}{Semi.}\end{tabular}} 
	    & PReMVOS~\cite{premvos} &\checkmark &\checkmark &\checkmark & 84.9	& 88.6 & - \\
	    & OSVOS~\cite{osvos}       &\checkmark & & & 79.8	& 80.6 & -\\ 
	    & MSK~\cite{msk}         &\checkmark &\checkmark &\checkmark & 79.7	& 75.4 & - \\ 
	    & PML~\cite{pml}         &\checkmark & & & 75.5 & 79.3 & -\\ 
	    & SFL~\cite{segflow} &\checkmark &\checkmark & & 76.1 & 76.0 & -\\
	    & VPN~\cite{vpn} &\checkmark &\checkmark & & 70.2 & 65.5 & -\\ \hline
	    \multirow{9}{*}{\begin{tabular}[r]{@{}c@{}}\rotatebox{90}{Unsupervised}\end{tabular}}
		& FST~\cite{fst}      & &\checkmark & & 55.8 & 51.1& 69.2 \\ 
		& ELM~\cite{elm}      & &\checkmark & & 61.8 & 61.2 & - \\ 
		& SFL~\cite{segflow}  & &\checkmark & & 67.4 & 66.7 & -\\ 
		& LMP~\cite{mp}       & &\checkmark &\checkmark & 70.0 & 65.9 & 77.5 \\ 
		& FSEG~\cite{fusionseg}& &\checkmark & & 70.7 & 65.3  & - \\ 
		& LVO~\cite{gru-mp}   & &\checkmark &\checkmark & 75.9 & 72.1 & 77.8 \\ 
		& ARP~\cite{arp}      & &\checkmark & & 76.2 & 70.6 &  \\ 
		& PDB~\cite{pdb}      & & &\checkmark & 77.2 & 74.5 & \textbf{81.5} \\
		& MotAdapt~\cite{teacher-student} & &\checkmark &\checkmark & 77.2 & 77.4 & 79.0 \\
	    \hline
	    & AD-Net (multiple scales) & & & & \bf{79.4} & \bf{78.2} & 81.2 \\
	    & AD-Net + I.Prun. (ours) & & & & \bf{81.7} & \bf{80.5} & - \\
	    \hline
	\end{tabular}
	}
	\vspace{0.1cm}
	\caption{Performance on DAVIS-2016 validation set. FF: first-frame annotations; OF: optical flow; CRF: random conditional field.}
	\label{table:davis-2016}
\end{table}

In Table~\ref{table:davis-2016}, we evaluate AD-Net against state-of-the-art unsupervised VOS methods on the DAVIS public leader-board and also provide the performance of several popular semi-supervised methods as a term of reference.
AD-Net attains the highest performance among all unsupervised methods on the DAVIS validation set, while also performing very competitively on the FBMS test set.
In particular, on DAVIS we outperform the second-best method (MotAdapt~\cite{teacher-student}) by an absolute margin of $2.2\%$ in $\mathcal{J}$ and $0.8\%$ in $\mathcal{F}$ before applying the post-processing step of instance pruning.
After applying instance pruning as described earlier, AD-Net achieves the final performance of $81.7$ in $\mathcal{J}$ and $80.5$ in $\mathcal{F}$, leading the second-best method by $4.5$ and $3.1$ absolute points respectively.
Also, despite being unsupervised at inference time, AD-Net outperforms many semi-supervised methods which instead require to be initialised with a mask in the first frame.

After our proposed AD-Net, the second and third-best ranking methods are MotAdapt~\cite{teacher-student} and PDB~\cite{pdb}, which are particularly representative of two classes of methods.

PDB is representative of top-performing RNN-based methods.
Although, in theory, RNNs could model long-range time dependencies, in practice they are constrained to model relatively short sequences.
First, as the computational graph of an (unrolled) RNN grows in depth with the length of a video sequence, backpropagation is typically limited to a few time steps (\eg, $5$ in RGMP~\cite{rgmp}).
Such backpropagation cannot guarantee long-term dependency modelling~\cite{sutskever}.
Second, despite the gating and memory mechanisms adopted by LSTMs and GRUs, long propagation paths of gradients still cause exploding or vanishing gradients~\cite{bengio,difficultRNN}.

Conversely, MotAdapt is representative of top-performing methods that employ optical flow.
It consists of a two-stream architecture, which dedicates two network branches (trained jointly but with different parameters) to process RGB images and pre-computed optical flow fields.
The two-branch network is further fine-tuned at inference time, with pseudo-labels generated by a teacher network.
Although optical flow is an intuitive way to model inter-frame dependencies and aid segmentation, results in Tables~\ref{tab:AD-Net} and~\ref{table:davis-2016} demonstrate that simply developing a better appearance-based model can overshadow the benefits of a dedicated optical flow branch.
Moreover, the strategy of fine-tuning at inference time adopted by MotAdapt and many semi-supervised methods is a time-consuming process, taking many seconds up to minutes per video.
In contrast, AD-Net leverages a simpler architecture, which makes it fast at inference time.
Without instance pruning, it runs online and at $4$ frames per second on an NVIDIA TITAN X GPU, with frames at the original DAVIS resolution of $854{\times}480$.
Speed can be easily traded off at a small cost in performance, by using frames with lower resolution and/or a lighter architecture.

\begin{table}[t]
\begin{center}
\small
\resizebox{\columnwidth}{!}
{
\begin{tabular}{c|r|cc|cc|cc}
\specialrule{1pt}{1pt}{1pt}
 & \multirow{2}{*}{\begin{tabular}[c]{c@{}} Saliency\\ Methods\end{tabular}}
 & \multicolumn{2}{c|}{DAVIS} 
 & \multicolumn{2}{c|}{FBMS} 
 & \multicolumn{2}{c}{ViSal} \\ 
 && MAE $\downarrow$ & F $\uparrow$ & MAE $\downarrow$ & F $\uparrow$ & MAE $\downarrow$ & F $\uparrow$  \\ \hline
 \specialrule{1pt}{1pt}{1pt}
 \multirow{11}{*}{\begin{tabular}[l]{c@{}}\rotatebox{90}{Image}\end{tabular}}
 &Amulet \cite{amulet} & 0.082 & 69.9 & 0.110 & 72.5 & 0.032 & 89.4 \\
 &SRM \cite{srm}    & 0.039 & 77.9 & 0.071 & 77.6 & 0.028 & 89.0 \\
 &UCF \cite{ucf}    & 0.107 & 71.6 & 0.147 & 67.9 & 0.068 & 87.0 \\
 &DSS \cite{dss}    & 0.062 & 71.7 & 0.083 & 76.4 & 0.028 & 90.6 \\
 &MSR \cite{msr}    & 0.057 & 74.6 & 0.064 & 78.7 & 0.031 & 90.1 \\
 &NLDF \cite{nldf}  & 0.056 & 72.3 & 0.092 & 73.6 & \textbf{0.023} & \textbf{91.6} \\
 &DCL \cite{dcl}    & 0.070 & 63.1 & 0.089 & 72.6 & 0.035 & 86.9 \\
 &DHS \cite{dhs}    & \textbf{0.039} & 75.8 & 0.083 & 74.3 & 0.025 & 91.1 \\
 &ELD \cite{eld}    & 0.070 & 68.8 & 0.103 & 71.9 & 0.038 & 89.0 \\
 &KSR \cite{ksr}    & 0.077 & 60.1 & 0.101 & 64.9 & 0.063 & 82.6 \\
 &RFCN \cite{rfcn}   & 0.065 & 71.0 & 0.105 & 73.6 & 0.043 & 88.8 \\
\hline
 \multirow{7}{*}{\begin{tabular}[l]{c@{}}\rotatebox{90}{Video}\end{tabular}}
 &FGRNE \cite{fgrne}    & 0.043 & 78.6 & 0.083 & 77.9 & 0.040 & 85.0 \\
 &FCNS \cite{fcns}     & 0.053 & 72.9 & 0.100 & 73.5 & 0.041 & 87.7 \\
 &SGSP \cite{sgsp}     & 0.128 & 67.7 & 0.171 & 57.1 & 0.172 & 64.8 \\
 &GAFL \cite{visal}     & 0.091 & 57.8 & 0.150 & 55.1 & 0.099 & 72.6 \\
 &SAGE \cite{sage}     & 0.105 & 47.9 & 0.142 & 58.1 & 0.096 & 73.4 \\
 &STUW \cite{stuw}     & 0.098 & 69.2 & 0.143 & 52.8 & 0.132 & 67.1 \\
 &SP \cite{sp}         & 0.130 & 60.1 & 0.161 & 53.8 & 0.126 & 73.1 \\
\hline
&AD-Net (ours) & 0.044 & \textbf{80.8} & \textbf{0.064} & \textbf{81.2} & 0.030 & 90.4 \\
\hline
\end{tabular}
}
\end{center}
\caption{Salient object detection performance of AD-Net, compared against $18$ popular saliency prediction methods.}
\label{tab:saliency}
\end{table}

\begin{figure*}[t]
\centering
\includegraphics[width=0.99\textwidth]{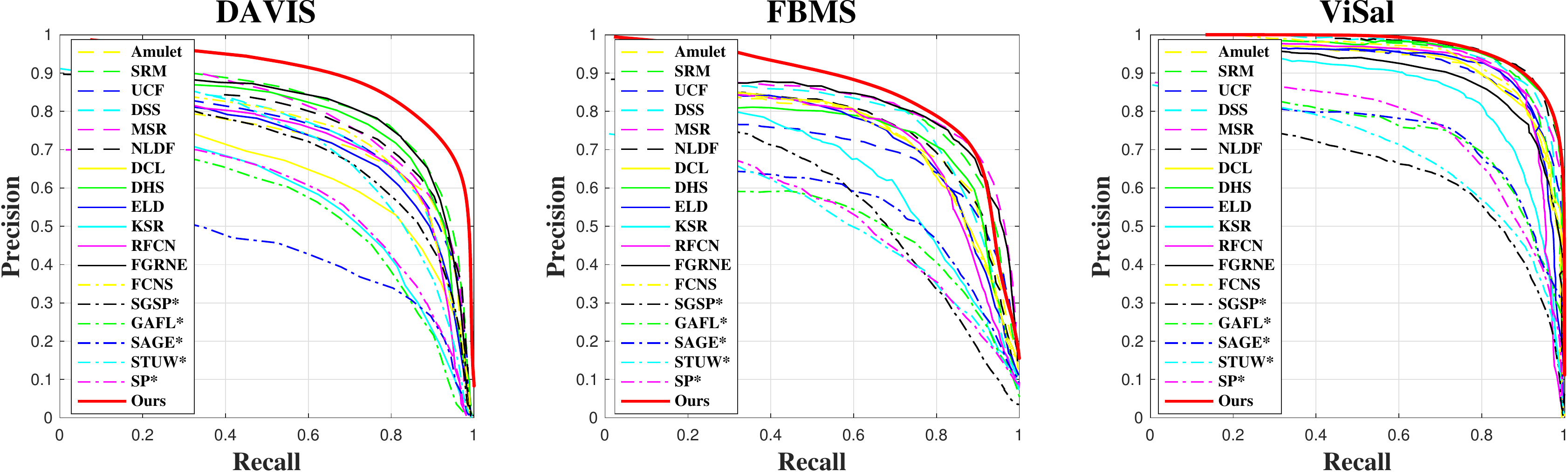}
\caption{AD-Net results with PR curves on the DAVIS, FBMS, and ViSal datasets.}
\label{fig:pr}  
\end{figure*}

The precision-recall analysis of AD-Net is presented in Figure~\ref{fig:pr}, where we demonstrate that our approach generally outperforms also existing salient object detection methods.
AD-Net achieves superior performance in all regions of the PR curve on the DAVIS validation set, maintaining significantly higher precision at all recall thresholds.
On the challenging FBMS test set, AD-Net maintains a clear advantage below the $90\%$ recall threshold.
On the ViSal dataset, it is noteworthy that nearly perfect precision is maintained up until the $60\%$ recall rate, which is higher than the other methods.
\begin{figure*}[t]
\centering
\includegraphics[width=0.99\textwidth]{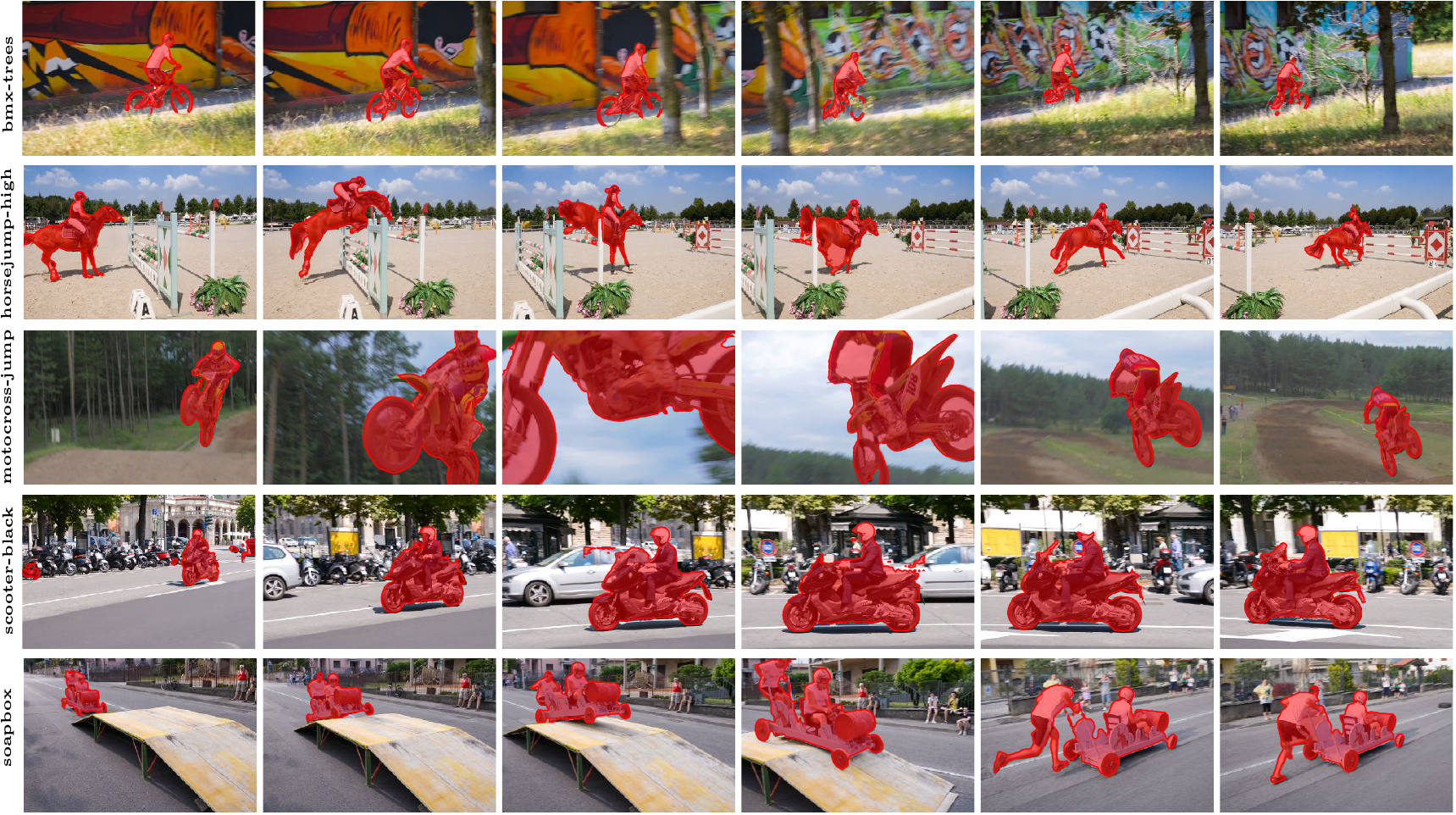}
\caption{Segmentation results on DAVIS-2016 validation set videos, obtained using our model without any online fine-tuning.}
\label{fig:segmentation}
\end{figure*}

\vspace{1ex}\noindent\textbf{Evaluation as video saliency}.
The definition of salient objects in a video for benchmarks like ViSal~\cite{visal} is very related to the one of ``foreground objects'' for benchmarks like DAVIS or FBMS (see Section~\ref{sec:introduction}).
Annotations in salient object detection datasets can vary from coarse annotations such as bounding boxes to fine-grained pixel-level real-valued scores, and sometimes even take the form of human eye fixations.
ViSal provides pixel-level annotations as binary labels, annotating large, moving objects as the foreground and everything else as the background.
Despite the many types of annotations, evaluation metrics are fairly standard and use pixel-level annotations either in a binarised form (PR curve and F-measure) or as normalised saliency scores between $0$ and $1$ (MAE), which are directly applicable to the scores produced by AD-Net.

As shown in Table~\ref{tab:saliency}, the proposed AD-Net improves the state of the art for both DAVIS and FBMS also for standard saliency scores, showing consistency with Table~\ref{table:davis-2016}.
The largest improvements lie in FBMS, where both MAE and F-measure significantly outperform previous records.
On DAVIS, F-measure is the highest among all methods with a significant leading margin.
On the ViSal dataset, AD-Net achieves best MAE (lower is better) among all video saliency models and obtains F-measure close to the overall best method.
Remarkably, despite not having trained for the task of saliency prediction, we outperform previous saliency methods under saliency metrics on DAVIS and FBMS, and achieve very competitive results on ViSal.


%% file: sections/conclusion.tex
\section{Conclusion} \label{sec: conclusion}
In this paper, we proposed Anchor Diffusion Network (AD-Net), a method for unsupervised video object segmentation based on non-local operations.
Instead of modelling temporal dependencies with recurrent connections or adopting pre-computed optical flow like contemporary work, we argue for a significantly simpler and more effective approach, which consists in establishing correspondences of pixel embeddings between a reference frame and the current one.
With this strategy, we can easily model long-term temporal dependencies at a low computational cost.
We show how, during inference, this procedure is able to suppress the background while preserving the foreground even when abrupt changes in appearance occur.
Quantitative evaluations across three standard benchmarks demonstrate the advantage of our proposed method on the task of unsupervised video object segmentation with respect to the state of the art.
Moreover, our method is also surprisingly competitive against the state of the art in semi-supervised video object segmentation and video saliency.

\vspace{1ex}\noindent\textbf{Acknowledgements}.
This work was supported by the ERC grant ERC-2012-AdG 321162-HELIOS, EPSRC grant Seebibyte EP/M013774/1, EPSRC/MURI grant EP/N019474/1, and Tencent.
We would also like to acknowledge the Royal Academy of Engineering and Five AI.

%% file: sections/supplementary.tex
\appendix

\section{Global Comparison} \label{sec:global}
\begin{table*}[htbp]
\begin{center}
\small
\resizebox{\textwidth}{!}
{
\begin{tabular}{r|ccccccccccccccccc}
\specialrule{1pt}{1pt}{1pt}
Measure &ADNet &MotAdapt\cite{teacher-student} &PDB\cite{pdb} &ARP\cite{arp} &LVO\cite{gru-mp} &FSEG\cite{fusionseg} &LMP\cite{mp} &SFL\cite{segflow} &TIS\cite{tis} &ELM\cite{elm} &FST\cite{fst} &CUT\cite{cut} &NLC\cite{nlc} &MSG\cite{msg} &KEY\cite{key} &CVOS\cite{causal} &TRC\cite{trc} \\ \hline
\rowcolor{rowblue}$\mathcal{J}$ Mean $\uparrow$ &\textbf{81.7} &77.2 &77.2 &76.2 &75.9 &70.7 &70.0 &67.4 &62.6 &61.8 &55.8 &55.2 &55.1 &53.3 &49.8 &48.2 &47.3 \\
$\mathcal{J}$ Recall $\uparrow$ &90.9 &87.8 &90.1 &\textbf{91.1} &89.1 &83.5 &85.0 &81.4 &80.3 &67.2 &64.9 &57.5 &55.8 &61.6 &59.1 &54.0 &49.3 \\
\rowcolor{rowblue}$\mathcal{J}$ Decay $\downarrow$ &2.2 &5.0 &0.9 &7.0 &\textbf{0.0} &1.5 &1.3 &6.2 &7.1 &9.8 &\textbf{0.0} &2.2 &12.6 &2.4 &14.1 &10.5 &8.3 \\ \hline
$\mathcal{F}$ Mean $\uparrow$ &\textbf{80.5} &77.4 &74.5 &70.6 &72.1 &65.3 &65.9 &66.7 &59.6 &61.2 &51.1 &55.2 &52.3 &50.8 &42.7 &44.7 &44.1 \\
\rowcolor{rowblue}$\mathcal{F}$ Recall $\uparrow$ &\textbf{85.1} &84.4 &84.4 &83.5 &83.4 &73.8 &79.2 &77.1 &74.5 &65.4 &51.6 &61.0 &51.9 &60.0 &37.5 &52.6 &43.6 \\
$\mathcal{F}$ Decay $\downarrow$ &0.6 &3.3 &\textbf{-0.2} &7.9 &1.3 &1.8 &2.5 &5.1 &6.4 &8.8 &2.9 &3.4 &11.4 &5.1 &10.6 &11.7 &12.9 \\ \hline
\rowcolor{rowblue}$\mathcal{T}$ (GT $8.8$) $\downarrow$ &36.9 &27.9 &29.1 &39.3 &26.5 &32.8 &57.2 &28.2 &33.6 &25.1 &36.6 &27.7 &42.5 &30.1 &26.9 &\textbf{25.0} &39.1 \\
\specialrule{1pt}{1pt}{1pt}
\end{tabular}
}
\end{center}
\caption{Detailed evaluation results on the DAVIS 2016 validation set. We analyse region similarity $\mathcal{J}$, contour accuracy $\mathcal{F}$, and temporal stability $\mathcal{T}$ in terms of mean, recall, and decay, and compare with state-of-the-art methods from the DAVIS 2016 leaderboard.}
\label{tab:overall}
\end{table*}
Table~\ref{tab:overall} includes all metrics reported in the official DAVIS 2016 benchmark~\cite{davis}.
Our method outperforms competing methods in the main evaluation metrics of mean region similarity $\mathcal{J}$ and mean contour accuracy $\mathcal{F}$.
The small decay measure for both $\mathcal{J}$ and $\mathcal{F}$ shows AD-Net's long-term benefits on performance.

\section{Per-sequence Comparison} \label{sec:per-seq}
\begin{figure*}[h!]
\centering
\includegraphics[width=0.995\textwidth]{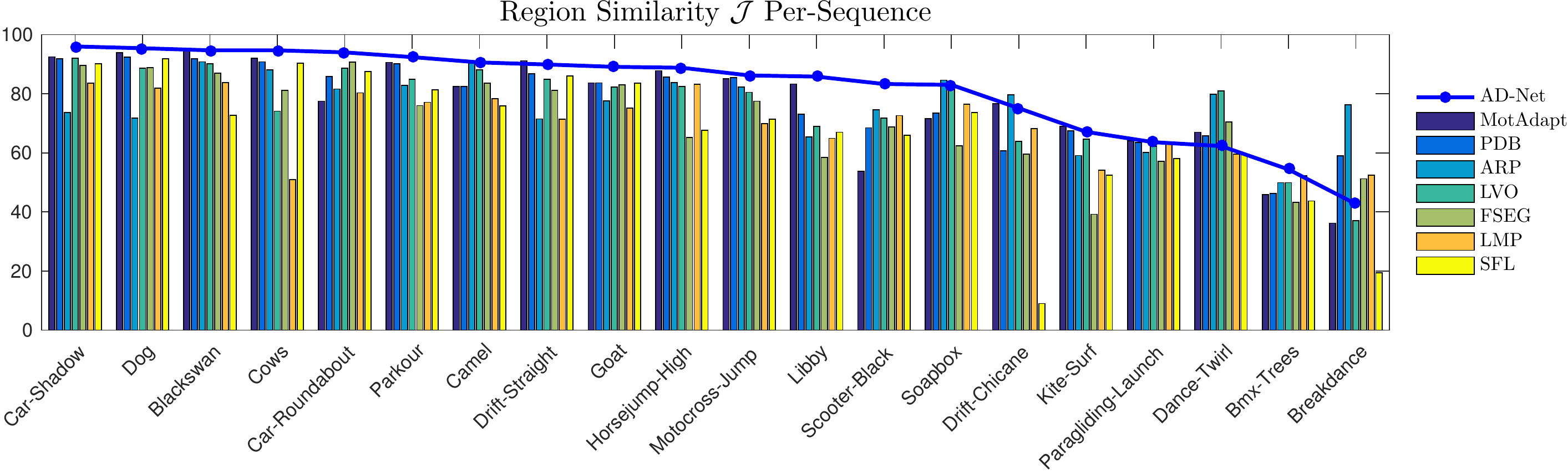}
\caption{Per-sequence results of mean region similarity $\mathcal{J}$ against top $7$ methods on the public leaderboard of DAVIS 2016. The blue line indicates AD-Net, while bars represent other methods. Sequences are organised in descending order of the performance of our method.}
\label{fig:j-seq}
\end{figure*}

\begin{figure*}[h!]
\centering
\includegraphics[width=0.995\textwidth]{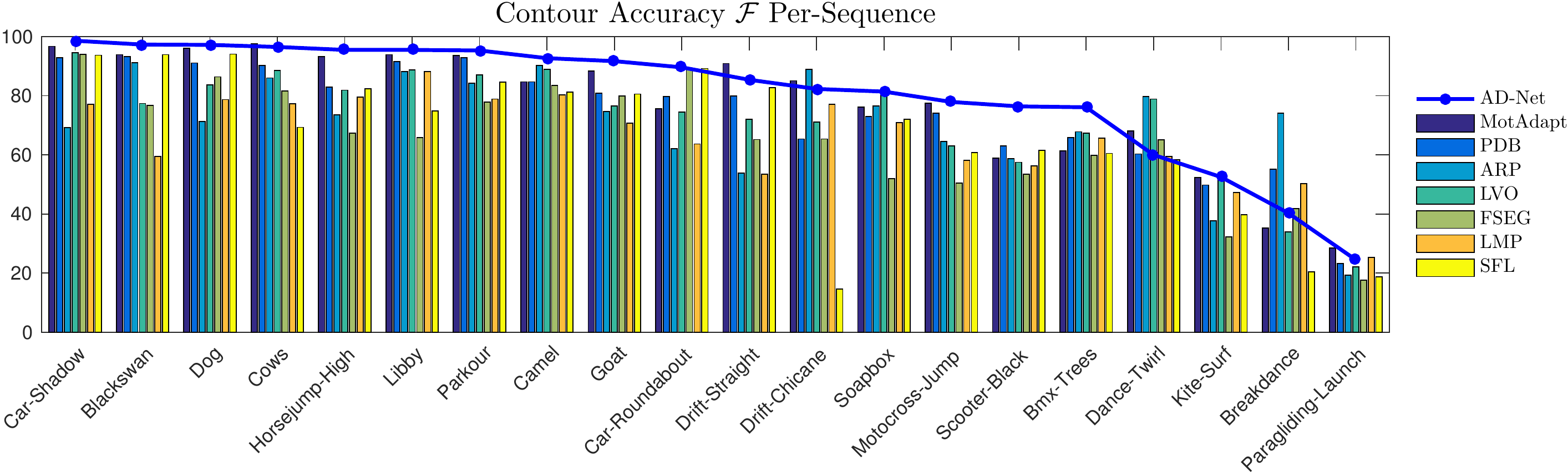}
\caption{Per-sequence results of mean contour accuracy $\mathcal{F}$ against top $7$ methods on the public leaderboard of DAVIS 2016. The blue line indicates AD-Net, while bars represent other methods. Sequences are organised in descending order of the performance of our method.}
\label{fig:f-seq}
\end{figure*}

Figures~\ref{fig:j-seq} and~\ref{fig:f-seq} compare the per-sequence $\mathcal{J}$ and $\mathcal{F}$ of AD-Net against the top seven competing methods on the leaderboard.
Our method performs well on videos presenting a variety of challenges, such as appearance change (Car-Shadow, Parkour), cluttered background (Car-Roundabout, Scooter-Black), occlusion (Libby, Bmx-Trees), fast motion (Bmx-Trees, Dog, Parkour),~\etc.

\begin{algorithm}[htb]
   \caption{Instance Pruning}
   \label{alg:2}
\begin{algorithmic}
   \STATE {\bfseries Input:} original masks $X=[x_0,...,x_{N-1}]$, bounding boxes/instance masks $E=[E_0,...,E_{M-1}]$, for $N$ frames and $M$ total instances
   \STATE {\bfseries Output:} refined masks $X'$
   \STATE  $size\_low \leftarrow Area(Sort(E)[-N])$
   \STATE $T \leftarrow SmallStatic(E, 0.6, 0.5N,size\_low)$
   
   \FOR{$t=1$ {\bfseries to} $N$}
   \STATE Let $b_t$ be instances on frame $t$ from $E$
   \STATE $F \leftarrow GetPruningMask(x_t, T, size\_low)$
   \STATE $x_t \leftarrow x_t\odot F$
   \ENDFOR
   \STATE \textbf{return} $X$
   
   \FUNCTION{$SmallStatic(b, iou, support, size)$}
   \STATE $sm\_stat\_instances \leftarrow \emptyset$
   \FOR{$b_i$ in $b$}
     \FOR{$b_j$ in $b$}
     \STATE $count \leftarrow 0$
     \IF{$IoU(b_i,b_j)>iou$}
     \STATE $count \leftarrow count + 1$
     \ENDIF
     \ENDFOR
     \IF{$count>support$ and $Size(b_i)<size$}
     \STATE Add $b_i$ to $sm\_stat\_instances$
     \ENDIF
   \ENDFOR
   \STATE \textbf{return} $sm\_stat\_instances$
   \ENDFUNCTION
   
   \FUNCTION{$GetPruningMask(x_t, T, s)$}
   \STATE $pruning\_mask \leftarrow \emptyset$, $target\_size \leftarrow -\infty$   
   \STATE $T_t \leftarrow Sort(T_t, descending)$
   \IF{$Size(T_t[0])>s$ \& $Len(T_t)>0$ \& $Size(T_t[0])>2Size(T_t[1])$}
   \STATE $target\_size \leftarrow Size(T_t[0])$
   \ENDIF
   \FOR{all $T_t[i]$ in $T_t$}
   \IF{$Size(T_t[i])<\frac{target\_size}{3}$}
   \STATE $pruning\_mask \leftarrow pruning\_mask \cup T_t[i]$
   \ENDIF
   \ENDFOR
   \STATE \textbf{return} $pruning\_mask$
   \ENDFUNCTION
\end{algorithmic}
\end{algorithm}

\section{Qualitative Analysis on FBMS and ViSal} \label{sec:qualitative}
\begin{figure*}[t]
\centering
\includegraphics[width=\textwidth]{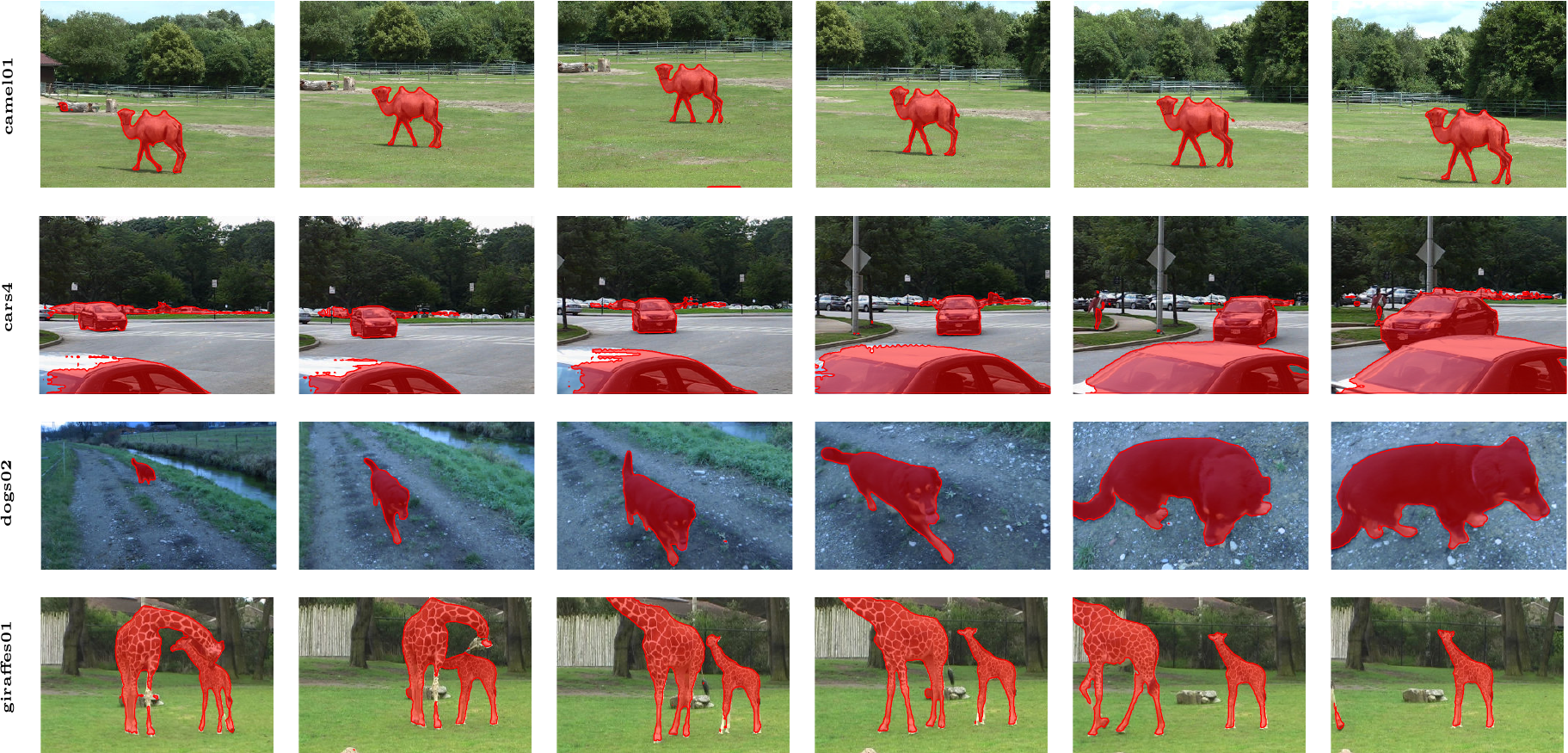}\\
\hspace{0.1cm}
\includegraphics[width=\textwidth]{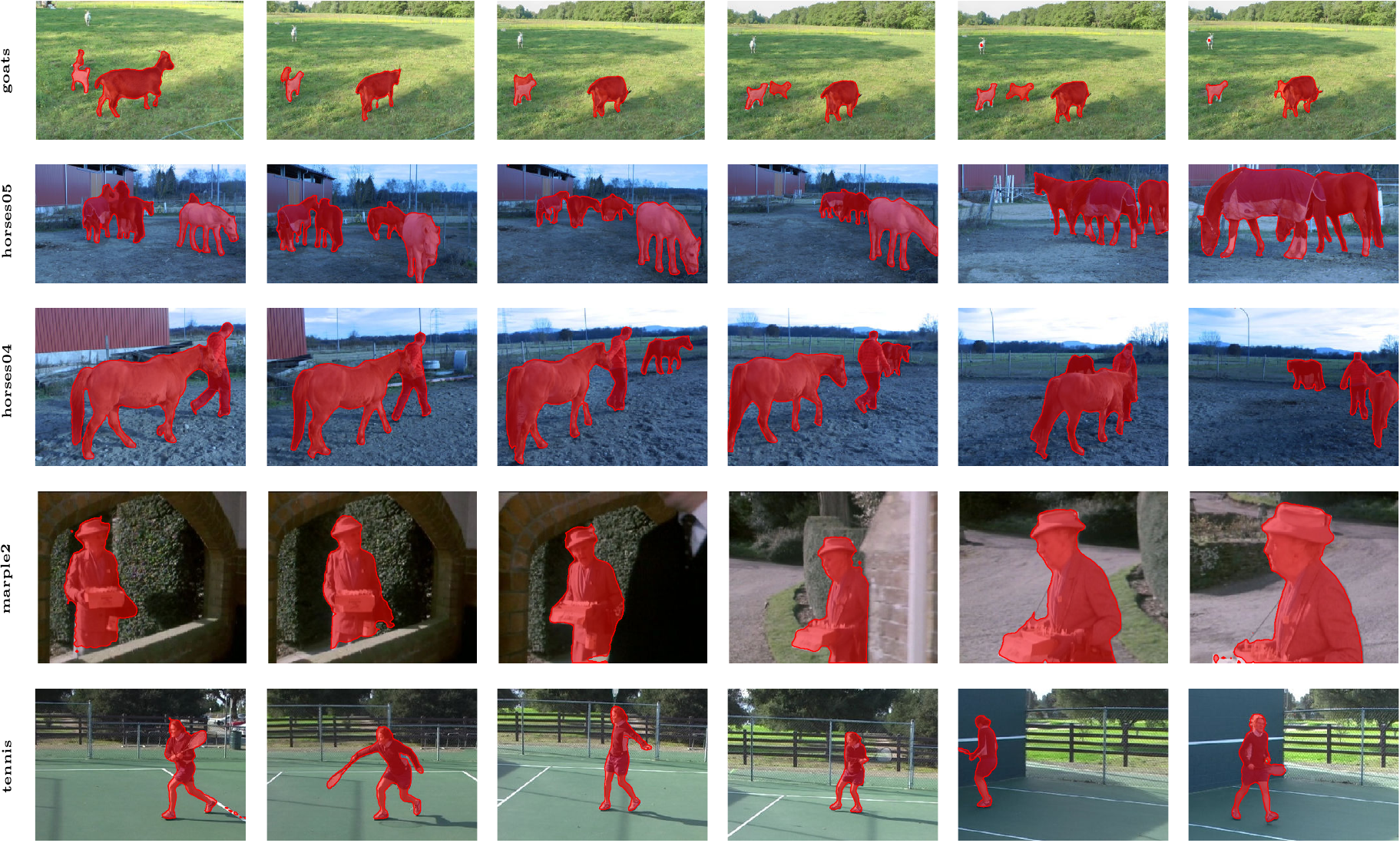}
\caption{Segmentation results on challenging videos from FBMS without fine-tuning.}
\label{fig:fbms}  
\end{figure*}

\begin{figure*}[t]
\centering
\includegraphics[width=\textwidth]{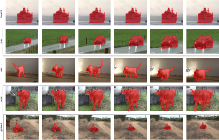}

\hspace*{1mm}

\includegraphics[width=\textwidth]{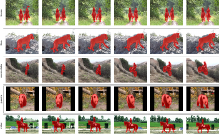}
\caption{Segmentation results on challenging videos from ViSal without fine-tuning.}
\label{fig:visal}
\end{figure*}

In Figures~\ref{fig:fbms} and~\ref{fig:visal}, we visualise segmentation results on videos from the test sets of FBMS~\cite{fbms} and ViSal~\cite{visal} respectively.
The model is trained only with the DAVIS 2016 training set.
We do not fine-tune it on the training set of FBMS or ViSal.

\section{Foreground Correspondence Analysis} \label{sec:correlation}
\begin{figure*}[tb]
\centering
\includegraphics[width=\textwidth]{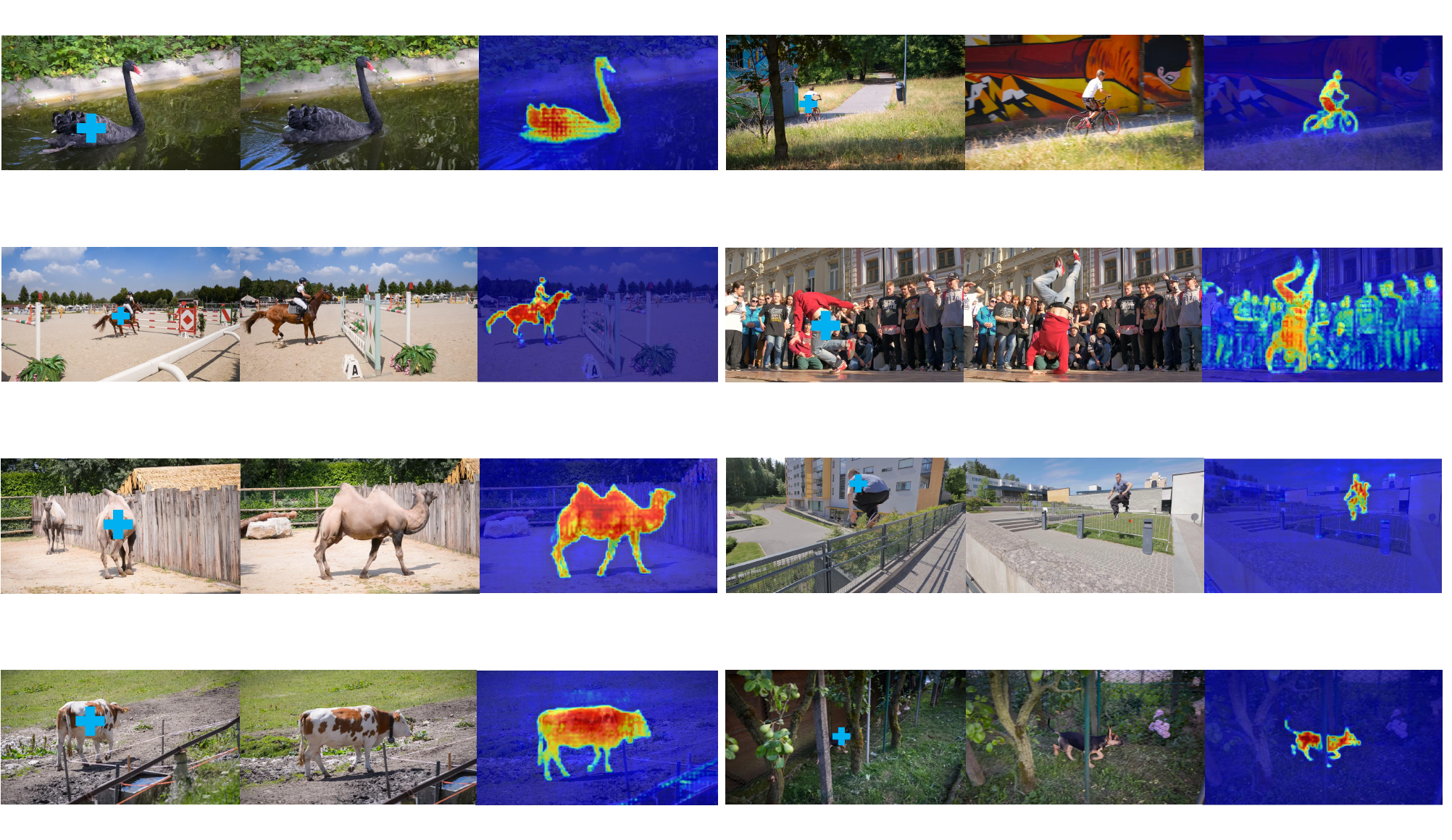}
\includegraphics[width=\textwidth]{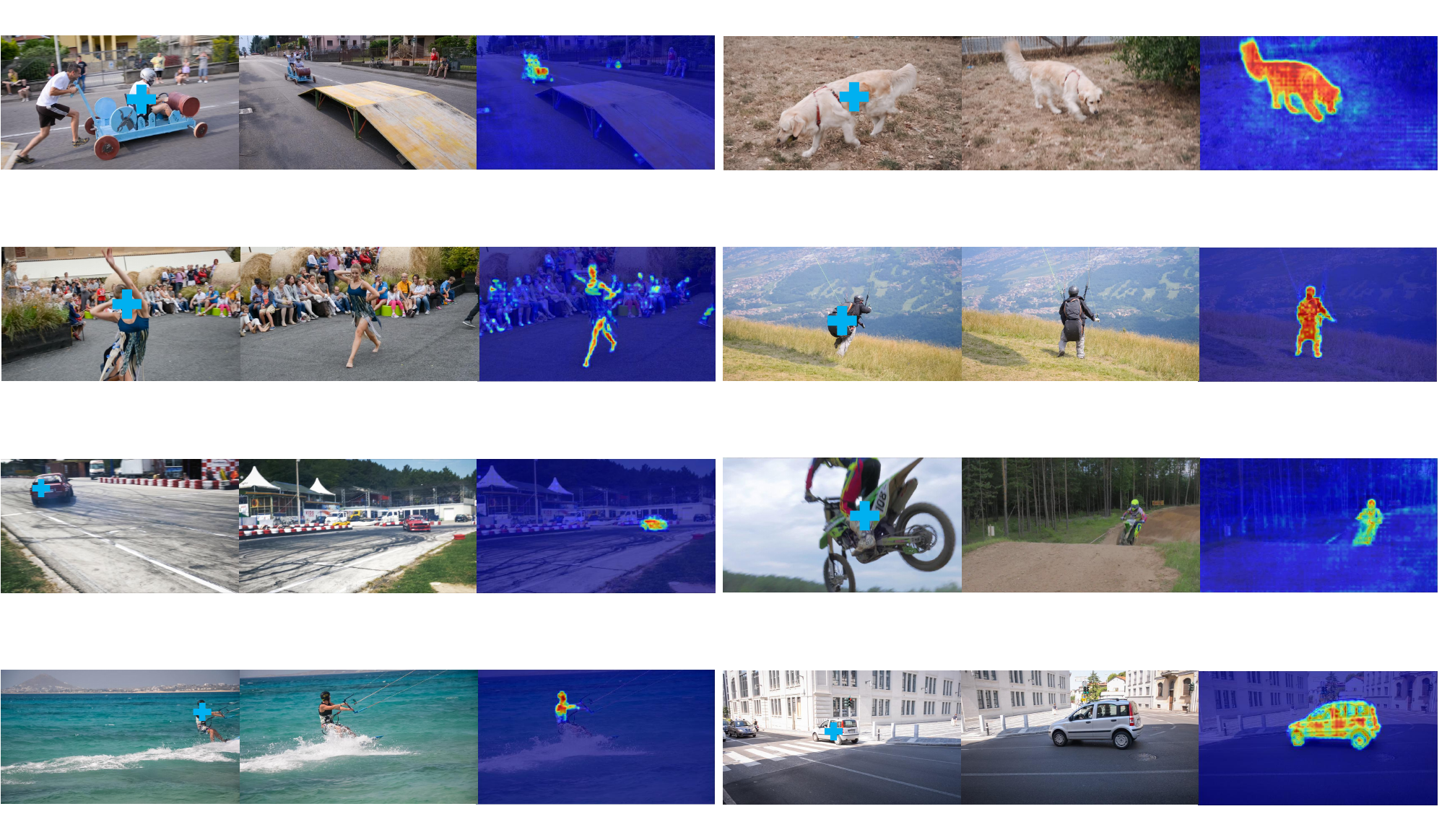}
\caption{Similarity scores of a foreground pixel on a distant frame with pixels in the anchor frame. The left, middle, and right images from each video illustrate, respectively, the target frame with the sampled foreground pixel (marked by the blue cross), the anchor frame, and similarities overlaid on the anchor frame.}
\label{fig:suppattention}
\end{figure*}

In Figure~\ref{fig:suppattention}, we visualise more examples of foreground pixel correspondences to pixels in the anchor frame.
Most pixels are randomly selected from the foreground area on the last frame of the video (except when foreground becomes too small in the last frame, in which case another frame is randomly chosen).

\section{Instance Pruning} \label{sec:instance-pruning}
Algorithm~\ref{alg:2} details the instance pruning procedure.
First, $SmallStatic$ returns a set of bounding boxes and the corresponding instance masks that represent small and nearly static instances.
Then, $GetPruningMask$ takes these instances and the original masks as inputs, and generates a pruning mask per frame, which incorporates all small and static instances that are much smaller than the largest instance in the current frame.
Finally, each input mask is multiplied element-wise with the corresponding pruning mask to output the final predictions.